\documentclass[9pt,conference]{IEEEtran}
\usepackage{amssymb,amsthm,amsmath,array}
\usepackage{graphicx}
\usepackage[caption=false,font=footnotesize]{subfig}
\usepackage{xspace}
\usepackage{stmaryrd}
\usepackage{url}
\usepackage{xcolor}
\usepackage{mathtools}
\usepackage{float}
\usepackage{textcomp}
\usepackage{adjustbox}
\usepackage{booktabs}
\usepackage[numbers]{natbib}

\begin{document}
\title{Enhancing Underwater Object Detection through Spatio-Temporal Analysis and Spatial Attention Networks}

\author{Sai Likhith Karri and Ansh Saxena}

\maketitle
\begin{abstract}
    This study examines the effectiveness of spatio-temporal modeling and the integration of spatial attention mechanisms in deep learning models for underwater object detection. Specifically, in the first phase, the performance of temporal-enhanced YOLOv5 variant T-YOLOv5 is evaluated, in comparison with the standard YOLOv5. For the second phase, an augmented version of T-YOLOv5 is developed, through the addition of a Convolutional Block Attention Module (CBAM). By examining the effectiveness of the already pre-existing YOLOv5 and T-YOLOv5 models and of the newly developed T-YOLOv5 with CBAM. With CBAM, the research highlights how temporal modeling improves detection accuracy in dynamic marine environments, particularly under conditions of sudden movements, partial occlusions, and gradual motion. The testing results showed that YOLOv5 achieved a mAP@50-95 of 0.563, while T-YOLOv5 and T-YOLOv5 with CBAM outperformed with mAP@50-95 scores of 0.813 and 0.811, respectively, highlighting their superior accuracy and generalization in detecting complex objects. The findings demonstrate that T-YOLOv5 significantly enhances detection reliability compared to the standard model, while T-YOLOv5 with CBAM further improves performance in challenging scenarios, although there is a loss of accuracy when it comes to simpler scenarios.
\end{abstract}

\section{Introduction}
The ocean is considered to be one of the largest locations on the planet for a variety and multitude of the resources on Earth. As marine exploration and resource extraction continue to expand, there is an increasing demand for advanced technologies to monitor and analyze underwater environments. Effective object detection is crucial for such applications, and, as a result, with growing artificial intelligence technologies, resource detection has taken major steps over time \cite{sun2024evolution, wang2023survey, jian2024underwater}. However, the unique challenges of underwater settings often hinder the performance of conventional object detection models. 

Addressing these challenges is essential for ensuring more accurate and reliable monitoring of marine ecosystems, especially as the need for real-time analysis grows. Whether it’s tracking marine wildlife, assessing the impact of resource extraction, or enhancing underwater robotics for exploration, reliable object detection can significantly improve operational efficiency and safety. Without the ability to accurately detect and track objects in such dynamic and complex environments, efforts to preserve marine biodiversity, manage resources, and develop autonomous underwater systems will be limited. Therefore, improving object detection technology for underwater settings is crucial not only for advancing scientific research but also for the sustainable management of the planet’s oceanic resources.

As aforementioned, the primary issue lies in the limitations of current object detection models when applied to underwater environments. Conventional methods struggle to maintain detection accuracy due to challenges such as blurred imagery, light scattering, and object occlusion, which are particularly prominent in underwater settings \cite{li2024novel, chen2023realtime}. These environmental factors compromise the ability of models to detect objects, especially in dynamic scenes where objects move abruptly or are partially obscured. In recent years, advances in deep learning models like You Only Look Once (YOLO) have demonstrated significant promise for object detection across various domains, including underwater environments. However, traditional models like YOLO, while promising  in UOD and effective in many domains, are not equipped to handle the low-contrast, blurry, and fluctuating visual conditions typical in underwater scenarios \cite{redmon2016yolo}. As a result, current models often fail to offer the necessary precision and consistency, particularly in real-time applications such as marine monitoring and underwater robotics. The inability to consistently detect and track objects under these conditions hinders progress in marine conservation, resource management, and underwater exploration \cite{zhang2023autonomous}. Thus, there is a critical need to develop more specialized models that can address these specific challenges and enhance the reliability of object detection in underwater environments.

This research aims to enhance underwater object detection by combining spatio-temporal modeling and spatial attention mechanisms within the YOLOv5 framework. By integrating ConvLSTM into YOLOv5, the model gains the ability to capture temporal dependencies across frames, improving detection stability in dynamic underwater scenes. At the same time, the inclusion of the Convolutional Block Attention Module (CBAM) strengthens the model’s ability to focus on important spatial features, even in conditions with occlusion, blur, or low contrast. Together, these enhancements are designed to significantly improve precision, consistency, and overall detection performance in challenging underwater environments.

This research contributes a unified approach that combines spatio-temporal modeling and attention mechanisms to advance object detection in complex visual environments. By enhancing an existing real-time detection framework with temporal awareness and refined spatial focus, we demonstrate how integrating these two techniques can lead to more accurate and consistent results. While our focus is underwater detection, the broader impact of this work lies in showing how temporal context and attention modules, two techniques which have not been thoroughly researched within previous underwater detection research, can be effectively applied together to improve performance in any setting where visual conditions are dynamic, noisy, or unpredictable. 
\section{Related Works}
Spatio-temporal modeling and attention mechanisms have catalyzed significant advancements in object detection across a wide range of domains. Methods such as ConvLSTM and Transformer-based architectures effectively capture temporal dependencies across video frames, enhancing detection stability in dynamic and fast-changing scenes. Likewise, attention mechanisms like CBAM and Coordinate Attention improve accuracy by allowing models to focus on the most salient features within complex or cluttered environments. In the context of underwater object detection, earlier research has primarily relied on conventional strategies—including image enhancement techniques, handcrafted feature extraction, and deep convolutional networks—designed to mitigate challenges such as low visibility, motion blur, and occlusions. These approaches have led to notable improvements in robustness and reliability. However, the integration of spatio-temporal and attention-based techniques in underwater settings remains in its early stages, offering considerable potential for addressing the persistent difficulties posed by the underwater environment. In the following section, we will explore two key areas: the evolution of underwater object detection methods and the broader application of spatio-temporal modeling and attention mechanisms in general object detection.

YOLO-based models have become a staple in underwater object detection due to their efficiency and real-time processing capabilities. Many researchers have proposed targeted enhancements to improve their robustness in underwater environments. For instance, Liu and Pan (2023) developed TC-YOLO \cite{liu2023underwater}, which integrates a Transformer encoder and Coordinate Attention into YOLOv5, along with CLAHE for image enhancement and optimal transport for label assignment. This approach achieved state-of-the-art results on the RUIE2020 dataset. Similarly, Wang et al. (2023) proposed UWV-YOLOX \cite{pan2023uwv_yolox}, which embeds Coordinate Attention into the backbone, introduces a novel loss function, and applies frame-level optimization—reaching a mAP@0.5 of 89.0\% on the UVODD dataset, a 3.2\% improvement over the base YOLO model. The Yolo Underwater model, also by Wang et al., uses dilated deformable convolutions and a dual-branch occlusion attention mechanism to better detect small or occluded objects—making it a relevant strategy for our research to consider when dealing with cluttered underwater imagery.

Building on these foundations, more complex and specialized models have emerged. BG-YOLO, features a bidirectional-guided framework that combines a parallel enhancement branch and detection branch, with a feature-guided module enabling the enhancement branch to inform detection—improving accuracy without added inference-time cost. CEH-YOLO incorporates a high-order deformable attention (HDA) module, an enhanced spatial pyramid pooling-fast (ESPPF) block, and a composite detection (CD) module, achieving mAPs of 88.4\% and 87.7\% on DUO and UTDAC2020, respectively, while operating at 156 FPS. At the cutting edge SU-YOLO leverages Spiking Neural Networks (SNNs) with a spike-based denoising module and Separated Batch Normalization (SeBN). This model delivers 78.8\% mAP with just 6.97 million parameters and an ultra-low energy cost of 2.98 mJ—opening up new possibilities for energy-efficient underwater perception. These diverse advancements offer a rich foundation of ideas and architectures that our research can build upon, especially when exploring the integration of spatio-temporal modeling and attention mechanisms.

Beyond YOLO-based models, several alternative approaches have been developed to enhance underwater object detection, addressing challenges such as low visibility, motion blur, and occlusions. For instance, the Underwater-YCC model integrates the Convolutional Block Attention Module (CBAM) into the network's neck, employs Conv2Former for feature fusion, and utilizes Wise-IoU for bounding box regression, resulting in a 1.49\% improvement in mean Average Precision (mAP) over conventional models like YOLOv5 and Faster R-CNN. Similarly, the MarineNet architecture applies a dual-branch design to separately handle object detection and background suppression, combining region-based convolutional networks with enhanced spatial attention mechanisms to improve detection in complex underwater scenes. The ADOD model introduces a residual attention mechanism to enhance feature focus and reduce background noise, improving detection accuracy and adaptability. While these advancements demonstrate significant progress, they often lack the integration of spatio-temporal modeling and attention mechanisms, which could further enhance detection performance in dynamic and challenging underwater conditions.

Spatio-temporal analysis in object detection has significantly advanced with the integration of deep learning techniques that capture both spatial and temporal dependencies. Early approaches combined convolutional neural networks (CNNs) with recurrent neural networks (RNNs) to model temporal sequences, enabling the detection of moving objects across frames. More recent developments have introduced Transformer-based architectures, such as the Spatio-Temporal Object Traces Attention Detection Transformer (ST-DETR), which utilize self-attention mechanisms to capture long-range dependencies and improve detection accuracy in dynamic environments. Additionally, models like Spatiotemporal Sampling Networks (STSN) employ deformable convolutions across time to learn spatially sampled features from adjacent frames, enhancing detection performance in videos. These advancements have been applied in various domains, including aerial vehicle detection using UAV cameras, where spatiotemporal models have demonstrated improved multi-class vehicle detection capabilities. In underwater object detection, spatio-temporal techniques could greatly enhance the detection of moving targets, even in challenging conditions like low visibility and motion blur, by leveraging both temporal dynamics and attention mechanisms to focus on key features across frames, thus improving accuracy and robustness in dynamic aquatic environments.

Attention mechanisms have seen significant advancements in object detection by allowing models to focus on the most important regions in an image, improving both performance and efficiency. In addition to CBAM and CSFA, other notable attention mechanisms include the Squeeze-and-Excitation Networks (SENet), which adaptively recalibrate channel-wise feature responses by learning an attention map that assigns importance to each channel. Non-Local Networks enhance this by capturing long-range dependencies in the image, allowing the model to focus on distant parts of the image for better contextual understanding. Another advancement is Spatial Attention, which helps focus on important spatial regions by generating attention maps based on spatial features, enhancing the model’s ability to focus on key areas in cluttered scenes. Additionally, Transformer-based attention mechanisms, like those used in the Vision Transformer (ViT), use self-attention layers to capture relationships between all pixels in the image, improving detection in complex scenes with varying scales and occlusions. By integrating such attention mechanisms into underwater object detection models, they could help address specific challenges such as low visibility and dynamic conditions, enabling the model to selectively focus on key features, objects, and temporal cues, improving accuracy and robustness.

In conclusion, the related works discussed, such as the advancements in YOLO-based models (e.g., TC-YOLO by Liu and Pan, and UWV-YOLOX by Wang et al.) and alternative techniques like Underwater-YCC, have laid a solid foundation for underwater object detection [9]. 

\section{Hypotheses}
Given prior research on spatio-temporal modeling in object detection and the advancements in integrating attention mechanisms to improve contextual understanding, the objectives of this research are to empirically test the following hypotheses for improved underwater object detection:
\begin{itemize}
    \item \textbf{H1}: Implementing temporal modeling in a YOLOv5 framework improves the accuracy and consistency of object detection in underwater video sequences compared to the standard YOLOv5 model.
    \item \textbf{H2}: Integrating a Convolutional Block Attention Module (CBAM) within a temporal YOLOv5 model enhances detection precision.
    \item \textbf{H3}: A YOLOv5-based spatio-temporal model, enhanced with CBAM, can more effectively capture and differentiate underwater objects across frames, as evidenced by higher confidence scores and reduced false positives compared to a non-spatially attentive model.
\end{itemize}

\section{Methodology}

\subsection{Dataset + Preprocessing}

\begin{figure}[H]
    \centering
    \includegraphics[width=0.5\textwidth]{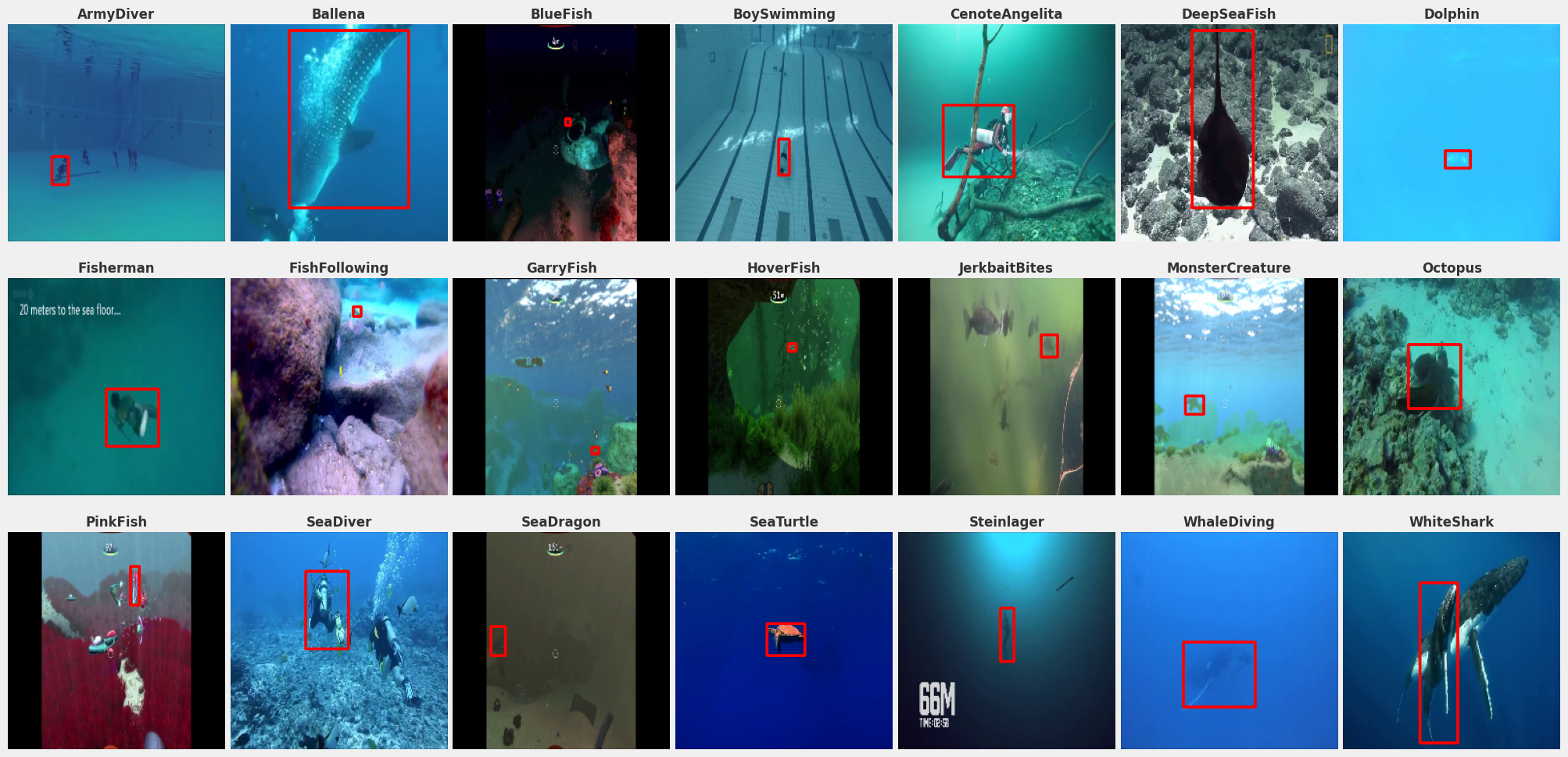}
    \caption{One sample image per class from the UOT32 Dataset}
    \label{fig:dataset_samples}
\end{figure}

The UOT32 Dataset [10], designed for underwater object tracking, contains 32 video sequences and 24,241 annotated frames, addressing challenges like low visibility, color distortions, and lighting inconsistencies in underwater environments. For consistency with established practices, we use individual frames from these videos as training data, allowing us to explore spatio-temporal modeling techniques by first training on static images and later incorporating temporal data [11, 12]. The dataset includes multiple object classes, such as fish, jellyfish, and plants, supporting the development of a robust detection framework. By integrating UOT32, we aim to refine YOLOv5 with spatio-temporal modeling, improving detection accuracy in dynamic underwater scenarios.

In the data preprocessing phase, we filtered out irrelevant whale images on beaches through manual inspection, focusing solely on underwater scenes for model relevance. Images were resized to 640x640 pixels to ensure compatibility with the YOLOv5 architecture, and bounding box annotations were converted to the YOLO format using normalized coordinates. For temporal data, we standardized video sequences to 100 frames, discarding excess frames to maintain consistency across the dataset. We then split the videos into training, validation, and test sets, ensuring no overlap and preserving temporal coherence. This approach allowed for efficient model training and unbiased performance evaluation, simulating real-world underwater conditions.

\subsection{Data Augmentations Incorporated In The T-YoloV5 Model}

\begin{figure}[H]
    \centering
    \begin{minipage}{0.24\textwidth}
        \centering
        \includegraphics[width=\textwidth]{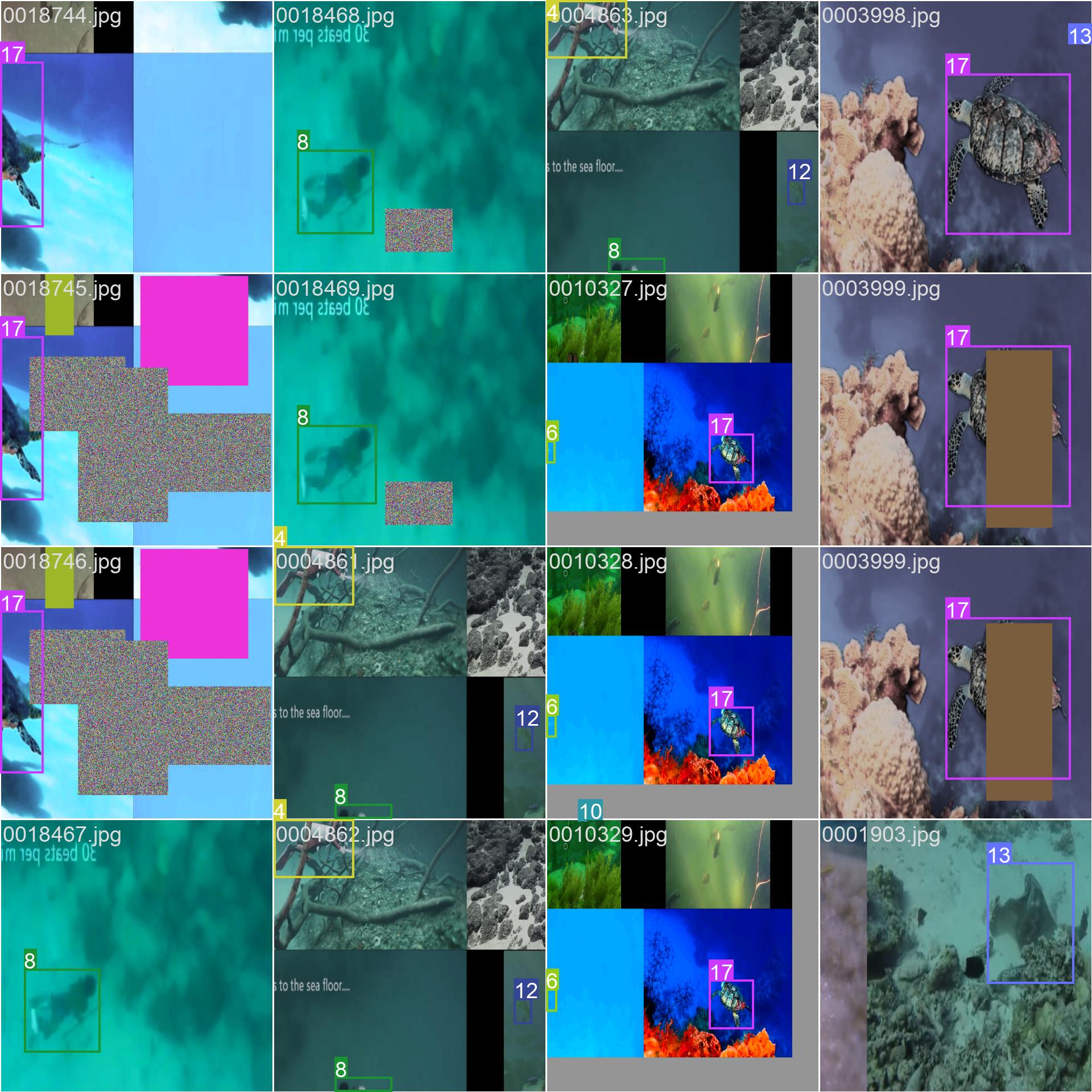}
    \end{minipage}%
    \hspace{0.02\textwidth}  
    \begin{minipage}{0.24\textwidth}
        \centering
        \includegraphics[width=\textwidth]{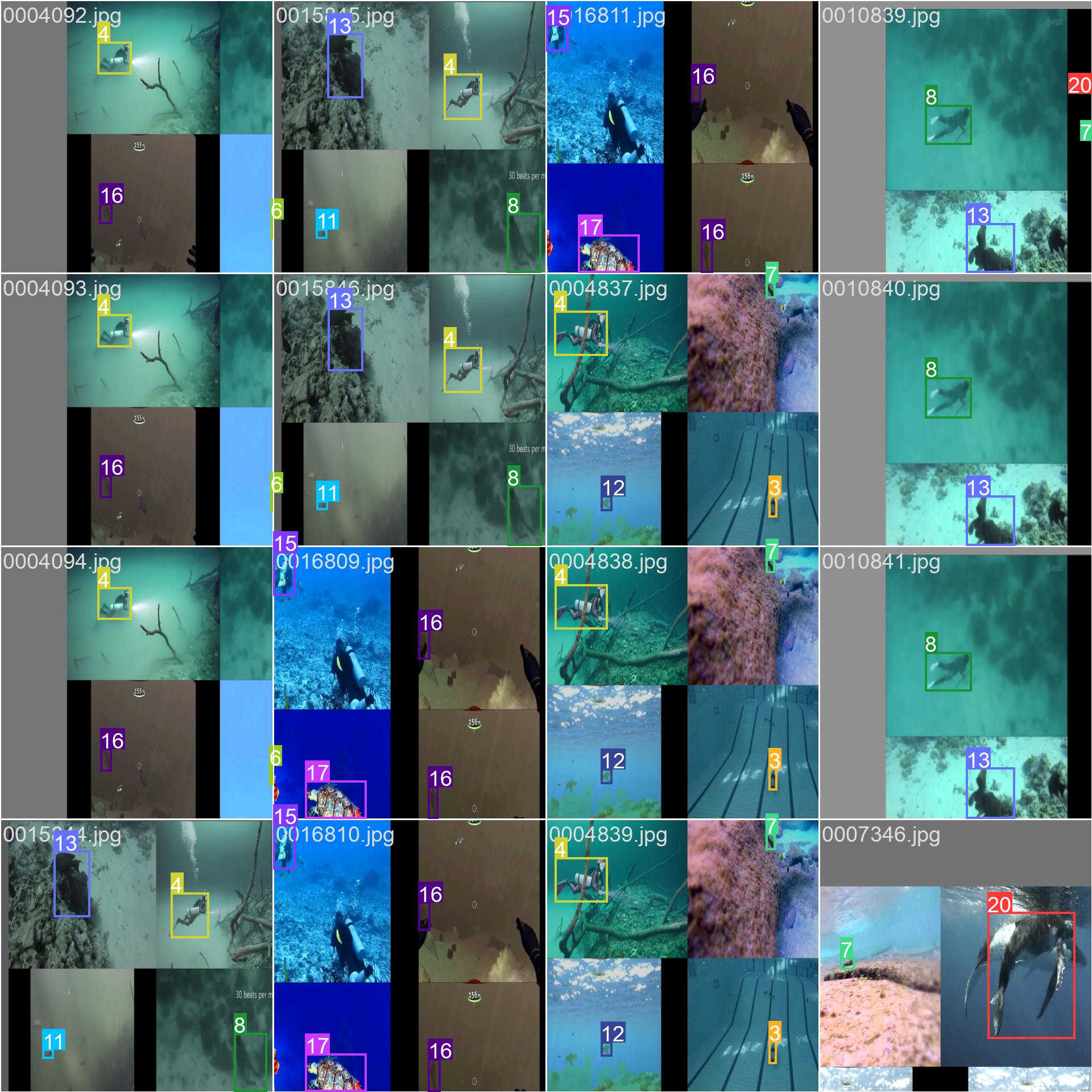}
    \end{minipage}
    \caption{Sample Images Utilized In Training Batches, After Augmentations.}
    \label{fig:model_architecture}
\end{figure}

Underwater object detection faces challenges like variable lighting, occlusions, and motion blur [13, 14]. To address these, several temporal data augmentations were incorporated into T-YOLOv5:
Temporal Mosaic Augmentation blends multiple scenes into a single frame, enhancing the model's adaptability to diverse underwater environments [15].
Temporal Mixup overlays frames from different scenes, helping the model recognize camouflaged objects in complex backgrounds.
Random Erasing simulates occlusions, encouraging the model to detect objects based on partial information.
Random Blur applies Gaussian blur to simulate motion or water-induced blur, preparing the model for dynamic, low-contrast conditions.
Gaussian Noise adds random noise, improving detection in low-light or murky environments.
These augmentations enhance the model's generalization, resilience to occlusions, and robustness in challenging underwater conditions.

\subsection{YOLOv5}

\begin{figure}[H]
    \centering
    \includegraphics[width=0.55\textwidth]{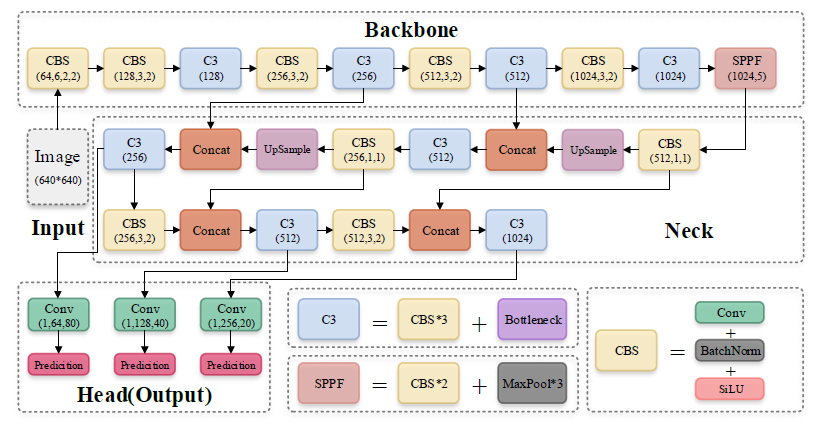}
    \caption{YOLOv5 Architecture Diagram}
    \label{fig:dataset_samples}
\end{figure}

YOLOv5 [16] has become a go-to model in underwater object detection due to its strong balance of speed and accuracy. Its modular architecture—comprising a backbone for feature extraction, a neck for multi-scale feature fusion, and a prediction head for detecting objects at different scales—makes it highly adaptable to the challenges of underwater scenes, such as variable lighting, small object sizes, and image noise. Many studies, like the UOD studies mentioned in our related works, have customized YOLOv5 to improve detection in underwater environments without compromising its efficiency [17, 18].

In our work, we also adopt YOLOv5 as the base model, leveraging its efficiency while extending its capabilities with spatio-temporal enhancements like ConvLSTM and attention modules. This allows us to build on a proven foundation and directly evaluate the impact of temporal modeling on detection performance in dynamic underwater settings.

\subsection{T-YOLOv5}

\begin{figure}[H]
    \centering
    \includegraphics[width=0.5\textwidth]{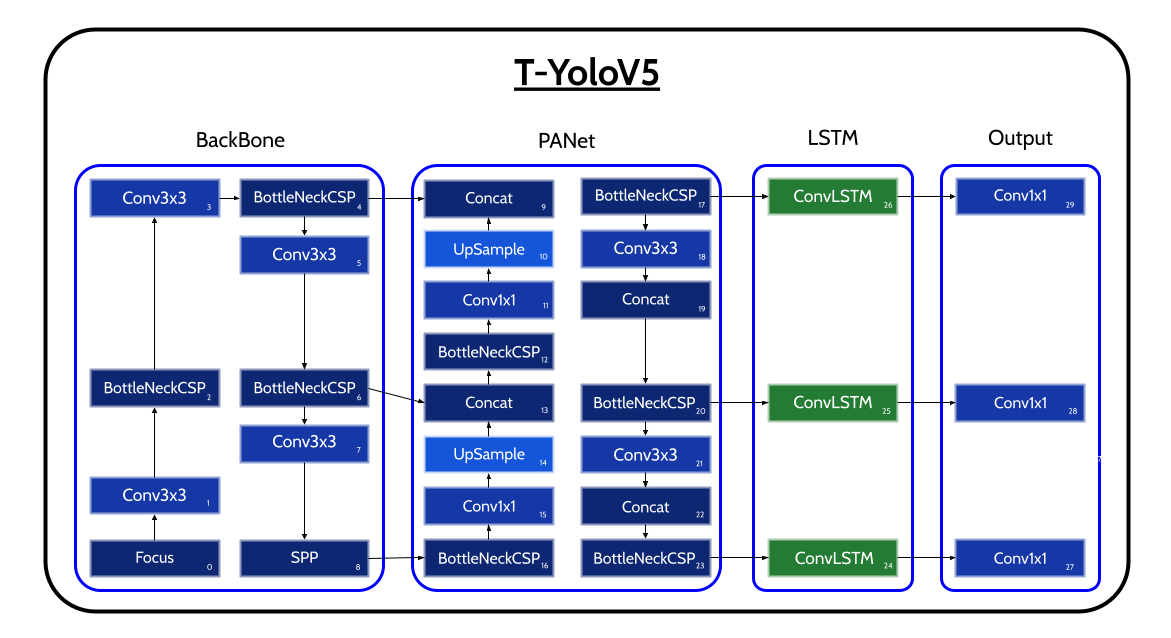}
    \caption{T-YOLOv5 Architecture Diagram}
    \label{fig:dataset_samples}
\end{figure}

T-YOLOv5 enhances the original YOLOv5 by incorporating temporal modeling through ConvLSTM, making it more effective for video-based object detection. Unlike the static-image focus of standard YOLOv5, T-YOLOv5 captures spatial-temporal patterns across frames, allowing it to track moving or partially obscured objects—crucial for underwater environments where motion, occlusions, and lighting distortions are common. ConvLSTM enables this by combining convolutional operations with LSTM's memory, preserving spatial structure across time.

\begin{figure}[H]
    \centering
    \includegraphics[width=0.5\textwidth]{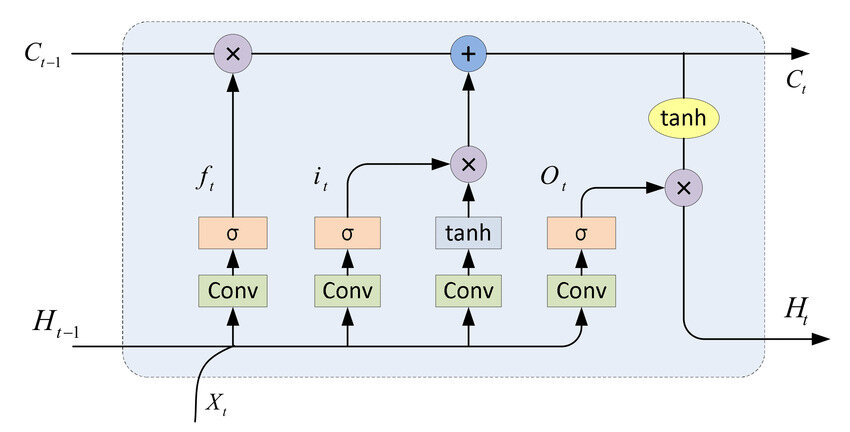}
    \caption{ConvLSTM Architecture Diagram}
    \label{fig:dataset_samples}
\end{figure}

For our project, we used a lightweight SConvLSTM version of T-YOLOv5 to accommodate limited processing power while still maintaining temporal context. We also benefited from an updated 2024 iteration of T-YOLOv5, which includes performance optimizations and compatibility improvements, allowing for a more precise application to underwater detection and a fair comparison to the latest YOLOv5 baseline.

\subsection{T-YOLOv5 With CBAM}

\begin{figure}[H]
    \centering
    \includegraphics[width=0.55\textwidth]{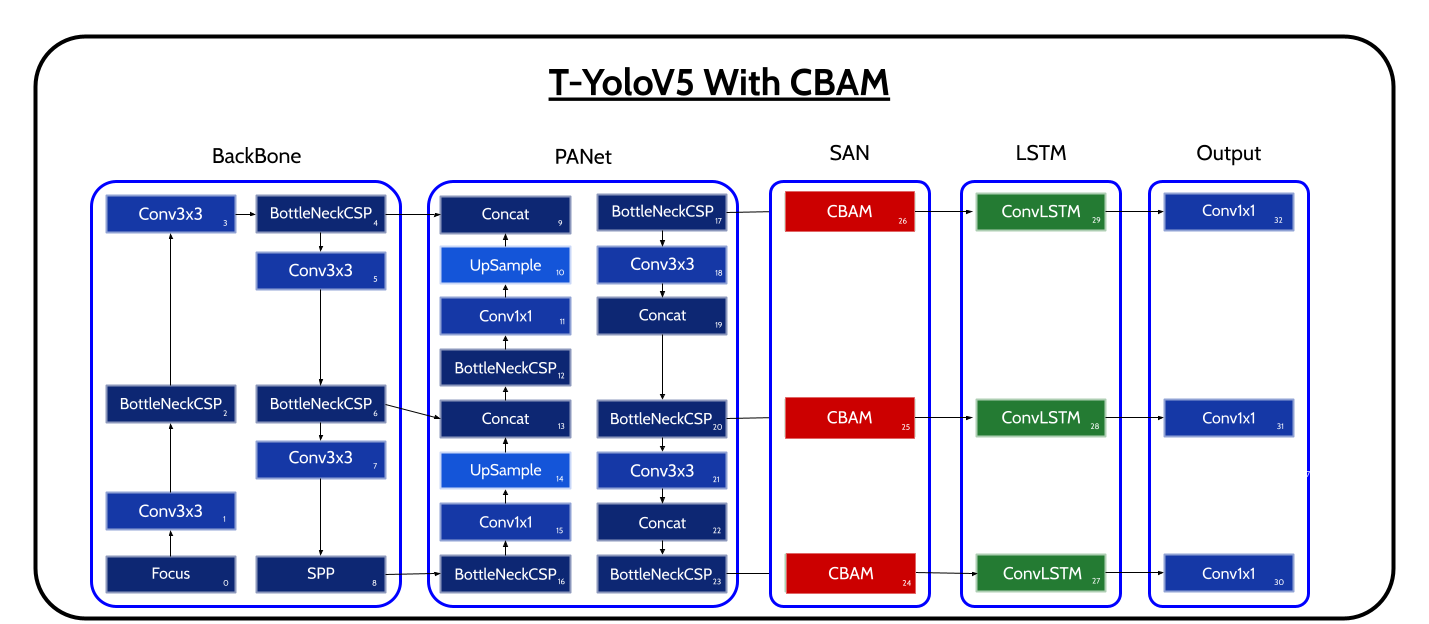}
    \caption{T-YOLOv5 with CBAM Architecture}
    \label{fig:dataset_samples}
\end{figure}

To improve detection in challenging underwater environments, we enhanced the T-YOLOv5 architecture by integrating the Convolutional Block Attention Module (CBAM) across key layers of the network. Underwater scenes often suffer from issues like low contrast, color shifts, and occlusions [19]. The core idea of spatial attention networks [20] like CBAM is to selectively focus on important areas in an image while suppressing less relevant regions, making them particularly effective in settings with background noise or clutter. CBAM addresses these challenges by refining feature maps through sequential channels and spatial attention [21, 22]. Spatial attention helps the model highlight important feature channels or determine where within an image to direct its focus, while channel attention informs the model what specific features are important across different channels of the feature map, improving object localization under poor visibility.

\begin{figure}[H]
    \centering
    \begin{minipage}{0.5\textwidth}
        \centering
        \includegraphics[width=\textwidth]{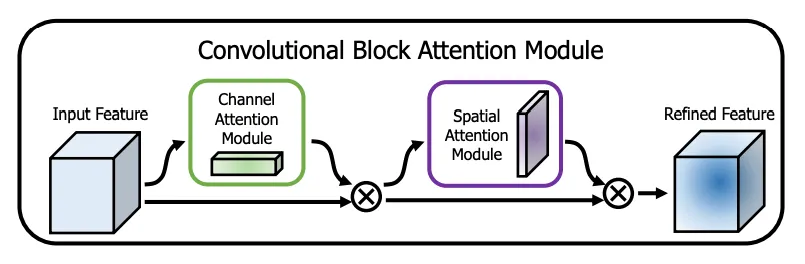}
    \end{minipage}%
    \hspace{0.0005\textwidth}  
    \begin{minipage}{0.5\textwidth}
        \centering
        \includegraphics[width=\textwidth]{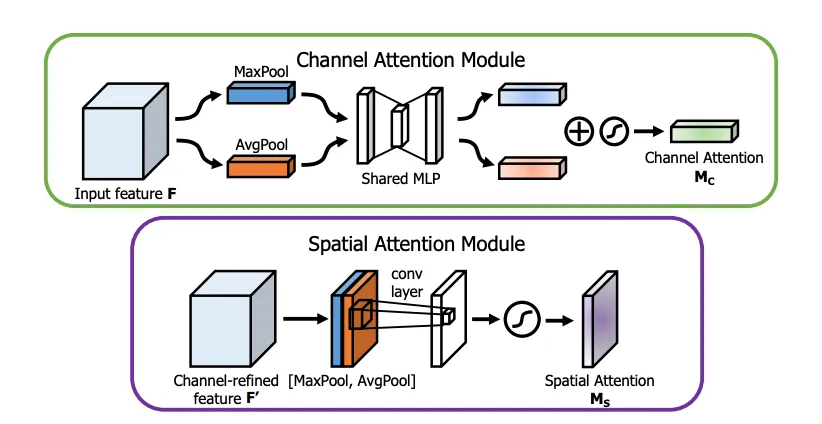}
    \end{minipage}
    \caption{CBAM Overview + Spatial and Channel Attention Structure}
    \label{fig:model_architecture}
\end{figure}

As shown in the architecture diagram, CBAM modules are placed after the PANet stage and before the ConvLSTM layer, allowing the network to adaptively focus on key features before spatio-temporal modeling. Each CBAM unit compresses and filters features via pooling operations and a lightweight MLP for channel attention, followed by convolutional refinement for spatial attention. This attention-enhanced output is then processed through ConvLSTM layers, capturing temporal consistency across frames before final predictions are made.

By embedding CBAM into T-YOLOv5’s multistage architecture—including the backbone, neck (PANet), and temporal layers—the model becomes more responsive to important textures and object boundaries while suppressing background noise. This dual attention mechanism significantly boosts performance in underwater object detection tasks where visual cues are often ambiguous or degraded.

\subsection{Training + Testing}
The standard YOLOv5 model was trained for 30 epochs with a 0.01 learning rate using pre-trained weights and a shuffled dataset split into training, validation, and testing sets. Validation metrics were recorded after each epoch for progress tracking. T-YOLOv5 was initialized with the best YOLOv5 weights and trained similarly for 30 epochs (learning rate 0.001, AdamW optimizer), using 3-frame sequences from 100-image video-like inputs to capture spatio-temporal features. The T-YOLOv5+CBAM model followed the same setup.

After training all three models, a final validation was conducted. Performance was evaluated using the test set, with metrics including mAP@50, mAP@50–90, precision, and recall to assess detection accuracy and generalization.

As a further test, the models were tested on all 32 UOT32 videos, generating frame-by-frame predictions with bounding boxes and class labels. A custom script was used to extract specific frame ranges, enabling focused analysis of detection performance in selected segments. This helped assess model adaptability to dynamic underwater environments, with corresponding diagrams shown in the results section.

\section{Results}
\subsection{Training}
The baseline YoloV5 showed a steady decrease in training loss and a gradual increase in precision and recall. However, its performance metrics plateaued at lower levels compared to the modified models. The validation loss was relatively consistent but higher than that of T-YoloV5 and T-YoloV5 with CBAM, indicating less effective generalization. The mAP scores for YoloV5 were the lowest among the three models, underscoring the benefits of using pre-trained weights and attention mechanisms for improved performance.

T-YoloV5, which began training with the best-performing weights from the baseline YoloV5, demonstrated a consistent decline in training losses that was smoother than that of YoloV5, indicating efficient use of the pre-trained weights and effective convergence. Precision and recall metrics showed strong upward trends, suggesting successful learning and better performance than YoloV5. Although validation loss showed occasional spikes, it remained lower than that of the baseline, pointing to improved generalization. The mAP scores for T-YoloV5 were also higher than those for YoloV5, highlighting the advantage of enhanced weights and architecture.

T-YOLOv5 with CBAM, initialized with optimal weights from T-YOLOv5, showed consistent decreases in training loss across box, object, and classification tasks, indicating effective convergence and enhanced feature extraction. Despite minor fluctuations, precision and recall consistently surpassed both the baseline and T-YOLOv5, underscoring CBAM’s impact on detection accuracy. While validation loss had occasional spikes, it generally declined, suggesting strong generalization. mAP scores at both 0.5 and 0.5:0.95 remained higher than in the other models, reinforcing CBAM’s benefit.
Initializing both T-YOLOv5 models with optimal weights accelerated training and improved performance. CBAM notably boosted feature representation, as seen in higher mAP and precision. Though validation loss varied at times, T-YOLOv5 with CBAM consistently outperformed its counterparts, while the baseline YOLOv5, though stable, lacked the advancements introduced by these enhancements.

\subsection{Validation \& Testing}

\begin{table}[!h]
    \centering
    \renewcommand{\arraystretch}{1.1} 
    \begin{minipage}{0.24\textwidth} 
        \centering
        \small
        \begin{adjustbox}{max width=\textwidth}
        \begin{tabular}{@{}l r c c c c@{}}
            \toprule
            \textbf{Class} & \textbf{Instances} & \textbf{P} & \textbf{R} & \textbf{mAP50} & \textbf{mAP50-95} \\
            \midrule
            all             & 4000 & 0.778 & 0.804 & 0.848 & 0.564 \\
            ArmyDiver       & 100  & 0.896 & 1.000 & 0.992 & 0.566 \\
            Ballena         & 200  & 0.996 & 1.000 & 0.992 & 0.563 \\
            BlueFish        & 200  & 0.865 & 0.960 & 0.973 & 0.485 \\
            BoySwimming     & 100  & 0.653 & 0.829 & 0.870 & 0.476 \\
            CenoteAngelita  & 200  & 0.994 & 1.000 & 0.991 & 0.699 \\
            DeepSeaFish     & 300  & 0.973 & 0.975 & 0.974 & 0.412 \\
            Dolphin         & 100  & 0.571 & 0.343 & 0.465 & 0.147 \\
            FishFollowing   & 300  & 0.743 & 0.921 & 0.874 & 0.458 \\
            Fisherman       & 100  & 0.944 & 0.970 & 0.966 & 0.509 \\
            GarryFish       & 200  & 1.000 & 0.270 & 0.881 & 0.333 \\
            HoverFish       & 400  & 0.992 & 0.992 & 0.992 & 0.644 \\
            MonsterCreature & 300  & 0.752 & 1.000 & 0.952 & 0.565 \\
            Octopus         & 600  & 0.901 & 0.833 & 0.889 & 0.476 \\
            PinkFish        & 100  & 0.830 & 0.821 & 0.921 & 0.482 \\
            SeaDragon       & 200  & 0.553 & 0.942 & 0.893 & 0.429 \\
            SeaTurtle       & 300  & 0.611 & 0.711 & 0.842 & 0.473 \\
            Steinlager      & 100  & 0.414 & 0.455 & 0.682 & 0.317 \\
            WhaleDiving     & 100  & 0.219 & 0.227 & 0.316 & 0.135 \\
            WhiteShark      & 100  & 0.993 & 1.000 & 0.994 & 0.676 \\
            \bottomrule
        \end{tabular}
        \end{adjustbox}
        \caption{Validation Metrics for YOLOv5}
        \label{table:performance_metrics_1}
    \end{minipage}%
    \hfill
    \begin{minipage}{0.24\textwidth} 
        \centering
        \small
        \begin{adjustbox}{max width=\textwidth}
        \begin{tabular}{@{}l r c c c c@{}}
            \toprule
            \textbf{Class} & \textbf{Instances} & \textbf{P} & \textbf{R} & \textbf{mAP50} & \textbf{mAP50-95} \\
            \midrule
            all             & 2100 & 0.866 & 0.803 & 0.851 & 0.563 \\
            ArmyDiver       & 300  & 0.00185 & 0.00186 & 0.0316 & 0.0273 \\
            BlueFish        & 400  & 0.984 & 0.895 & 0.961 & 0.585 \\
            BoySwimming     & 100  & 0.994 & 1.000 & 0.995 & 0.666 \\
            DeepSeaFish     & 200  & 0.972 & 0.693 & 0.941 & 0.472 \\
            Dolphin         & 200  & 0.989 & 1.000 & 0.995 & 0.678 \\
            FishFollowing   & 100  & 1.000 & 0.875 & 0.990 & 0.580 \\
            MonsterCreature & 100  & 0.995 & 1.000 & 0.995 & 0.832 \\
            Octopus         & 200  & 0.908 & 0.886 & 0.949 & 0.524 \\
            SeaDiver        & 100  & 0.997 & 1.000 & 0.995 & 0.546 \\
            SeaDragon       & 100  & 0.994 & 1.000 & 0.995 & 0.784 \\
            SeaTurtle       & 200  & 0.911 & 1.000 & 0.995 & 0.873 \\
            WhiteShark      & 100  & 0.642 & 0.280 & 0.373 & 0.193 \\
            \bottomrule
        \end{tabular}
        \end{adjustbox}
        \caption{Testing Metrics for YOLOv5}
        \label{table:performance_metrics_2}
    \end{minipage}
\end{table}

Validation and testing results of the YOLOv5 models highlight key performance trends. The standard YOLOv5 (157 layers, 15.9 GFLOPs) delivered solid baseline results, with a validation mAP@50 of 0.848, mAP@50-95 of 0.564, precision of 0.893, and recall of 0.822. It performed well on clear classes (e.g., ArmyDiver, Ballena, BoySwimming) but struggled with more ambiguous ones like GarryFish and PinkFish.

\begin{table}[H]
    \centering
    \renewcommand{\arraystretch}{1.1} 
    \begin{minipage}{0.24\textwidth} 
        \centering
        \small
        \begin{adjustbox}{max width=\textwidth}
        \begin{tabular}{@{}l r c c c c@{}}
        \toprule
        \textbf{Class} & \textbf{Instances} & \textbf{P} & \textbf{R} & \textbf{mAP50} & \textbf{mAP50-95} \\
        \midrule
        all             & 4000 & 0.788 & 0.804 & 0.868 & 0.464 \\
        ArmyDiver       & 100  & 0.902 & 1.000 & 0.995 & 0.565 \\
        Ballena         & 200  & 0.996 & 1.000 & 0.995 & 0.561 \\
        BlueFish        & 200  & 0.875 & 0.960 & 0.975 & 0.482 \\
        BoySwimming     & 100  & 0.658 & 0.829 & 0.873 & 0.470 \\
        CenoteAngelita  & 200  & 0.997 & 1.000 & 0.995 & 0.697 \\
        DeepSeaFish     & 300  & 0.979 & 0.980 & 0.972 & 0.417 \\
        Dolphin         & 100  & 0.574 & 0.340 & 0.465 & 0.143 \\
        FishFollowing   & 300  & 0.748 & 0.923 & 0.877 & 0.457 \\
        Fisherman       & 100  & 0.951 & 0.972 & 0.965 & 0.506 \\
        GarryFish       & 200  & 1.000 & 0.273 & 0.882 & 0.334 \\
        HoverFish       & 400  & 0.995 & 0.997 & 0.995 & 0.641 \\
        MonsterCreature & 300  & 0.754 & 1.000 & 0.951 & 0.563 \\
        Octopus         & 600  & 0.905 & 0.835 & 0.890 & 0.471 \\
        PinkFish        & 100  & 0.830 & 0.829 & 0.923 & 0.483 \\
        SeaDragon       & 200  & 0.558 & 0.940 & 0.898 & 0.428 \\
        SeaTurtle       & 300  & 0.612 & 0.713 & 0.845 & 0.477 \\
        Steinlager      & 100  & 0.417 & 0.457 & 0.687 & 0.312 \\
        WhaleDiving     & 100  & 0.220 & 0.230 & 0.319 & 0.136 \\
        WhiteShark      & 100  & 0.995 & 1.000 & 0.995 & 0.676 \\
        \bottomrule
        \end{tabular}
        \end{adjustbox}
        \caption{Validation Metrics for T-YOLOv5}
        \label{table:performance_metrics_1}
    \end{minipage}%
    \hfill
    \begin{minipage}{0.24\textwidth} 
        \centering
        \small
        \begin{adjustbox}{max width=\textwidth}
        \begin{tabular}{@{}l r c c c c@{}}
        \toprule
        \textbf{Class} & \textbf{Instances} & \textbf{P} & \textbf{R} & \textbf{mAP50} & \textbf{mAP50-95} \\
        \midrule
        all             & 2100 & 0.937 & 0.993 & 0.951 & 0.813 \\
        ArmyDiver       & 300  & 0.493 & 0.973 & 0.497 & 0.443 \\
        BlueFish        & 400  & 0.986 & 0.998 & 0.995 & 0.816 \\
        BoySwimming     & 100  & 0.990 & 1.000 & 0.995 & 0.925 \\
        DeepSeaFish     & 200  & 0.993 & 1.000 & 0.995 & 0.882 \\
        Dolphin         & 200  & 0.984 & 1.000 & 0.995 & 0.729 \\
        FishFollowing   & 100  & 0.954 & 0.990 & 0.994 & 0.701 \\
        MonsterCreature & 100  & 0.989 & 1.000 & 0.995 & 0.862 \\
        Octopus         & 200  & 0.947 & 0.955 & 0.969 & 0.802 \\
        SeaDiver        & 100  & 0.991 & 1.000 & 0.995 & 0.898 \\
        SeaDragon       & 100  & 0.990 & 1.000 & 0.995 & 0.884 \\
        SeaTurtle       & 200  & 0.941 & 1.000 & 0.995 & 0.934 \\
        WhiteShark      & 100  & 0.991 & 1.000 & 0.995 & 0.883 \\
        \bottomrule
        \end{tabular}
        \end{adjustbox}
        \caption{Testing Metrics for T-YOLOv5}
        \label{table:performance_metrics_2}
    \end{minipage}
\end{table}

T-YOLOv5 (170 layers, 61.2 GFLOPs) significantly improved detection accuracy, achieving a validation mAP@50 of 0.868 and mAP@50-95 of 0.464, with testing mAP@50 of 0.951 and mAP@50-95 of 0.813. Its high recall (0.993) and strong results on complex classes like BlueFish and SeaDragon underscore the benefits of temporal modeling.

\begin{table}[H]
    \centering
    \renewcommand{\arraystretch}{1.1} 
    \begin{minipage}{0.24\textwidth} 
        \centering
        \small
        \begin{adjustbox}{max width=\textwidth}
        \begin{tabular}{@{}l r c c c c@{}}
        \toprule
        \textbf{Class} & \textbf{Instances} & \textbf{P} & \textbf{R} & \textbf{mAP50} & \textbf{mAP50-95} \\
        \midrule
            all               & 4000 & 0.857 & 0.779 & 0.864 & 0.467 \\
            ArmyDiver         & 100  & 0.991 & 1.000 & 0.995 & 0.662 \\
            Ballena           & 200  & 0.970 & 1.000 & 0.995 & 0.551 \\
            BlueFish          & 200  & 0.954 & 0.870 & 0.973 & 0.512 \\
            BoySwimming       & 100  & 0.696 & 0.802 & 0.906 & 0.480 \\
            CenoteAngelita    & 200  & 0.997 & 1.000 & 0.995 & 0.683 \\
            DeepSeaFish       & 300  & 0.976 & 0.977 & 0.972 & 0.373 \\
            Dolphin           & 100  & 0.946 & 0.350 & 0.565 & 0.217 \\
            FishFollowing     & 300  & 0.775 & 0.910 & 0.873 & 0.426 \\
            Fisherman         & 100  & 0.935 & 0.940 & 0.960 & 0.512 \\
            GarryFish         & 200  & 0.936 & 0.221 & 0.688 & 0.261 \\
            HoverFish         & 400  & 0.990 & 0.991 & 0.994 & 0.662 \\
            MonsterCreature   & 300  & 0.669 & 0.810 & 0.915 & 0.553 \\
            Octopus           & 600  & 0.942 & 0.846 & 0.885 & 0.454 \\
            PinkFish          & 100  & 0.603 & 0.720 & 0.663 & 0.288 \\
            SeaDragon         & 200  & 0.893 & 0.833 & 0.907 & 0.485 \\
            SeaTurtle         & 300  & 0.853 & 0.831 & 0.951 & 0.485 \\
            Steinlager        & 100  & 0.932 & 0.470 & 0.768 & 0.451 \\
            WhaleDiving       & 100  & 0.225 & 0.230 & 0.423 & 0.157 \\
            WhiteShark        & 100  & 0.996 & 1.000 & 0.995 & 0.658 \\
            \bottomrule
        \end{tabular}
        \end{adjustbox}
        \caption{Validation Metrics for T-YOLOv5 With CBAM}
        \label{table:performance_metrics_val}
    \end{minipage}%
        \hfill
    \begin{minipage}{0.24\textwidth} 
        \centering
        \small
        \begin{adjustbox}{max width=\textwidth}
        \begin{tabular}{@{}l r c c c c@{}}
        \toprule
        \textbf{Class} & \textbf{Instances} & \textbf{P} & \textbf{R} & \textbf{mAP50} & \textbf{mAP50-95} \\
        \midrule
            all               & 2100 & 0.938 & 0.994 & 0.951 & 0.811 \\
            ArmyDiver         & 300  & 0.496 & 0.986 & 0.497 & 0.448 \\
            BlueFish          & 400  & 0.984 & 0.998 & 0.994 & 0.813 \\
            BoySwimming       & 100  & 0.991 & 1.000 & 0.995 & 0.905 \\
            DeepSeaFish       & 200  & 0.987 & 1.000 & 0.995 & 0.863 \\
            Dolphin           & 200  & 0.986 & 1.000 & 0.995 & 0.742 \\
            FishFollowing     & 100  & 0.969 & 0.990 & 0.994 & 0.702 \\
            MonsterCreature   & 100  & 0.991 & 1.000 & 0.995 & 0.863 \\
            Octopus           & 200  & 0.947 & 0.955 & 0.969 & 0.793 \\
            SeaDiver          & 100  & 0.991 & 1.000 & 0.995 & 0.898 \\
            SeaDragon         & 100  & 0.990 & 1.000 & 0.995 & 0.900 \\
            SeaTurtle         & 200  & 0.933 & 1.000 & 0.995 & 0.947 \\
            WhiteShark        & 100  & 0.991 & 1.000 & 0.995 & 0.861 \\
            \bottomrule
        \end{tabular}
        \end{adjustbox}
        \caption{Testing Metrics for T-YOLOv5 With CBAM}
        \label{table:performance_metrics_test}
    \end{minipage}
\end{table}

T-YOLOv5 with CBAM (207 layers, 61.3 GFLOPs) added attention mechanisms to refine feature focus. It maintained strong performance (validation mAP@50=0.864, mAP@50-95=0.467; testing mAP@50=0.951, mAP@50-95=0.811) but had slightly lower precision (0.857). It excelled in classes like HoverFish and SeaTurtle but showed inconsistencies on WhaleDiving and GarryFish, suggesting challenges with smaller or less distinct objects even with CBAM.

\textbf{General Trends and Best Model}

The progression from YOLOv5 to T-YOLOv5 and T-YOLOv5 with CBAM reveals several trends:

\begin{itemize}
    \item \textbf{Improved Recall and mAP:} Both temporal models outperform the baseline in accuracy and generalization.
    \item \textbf{Class-Specific Gains:} CBAM enhances detection of detailed or partially occluded objects.
    \item \textbf{Computational Trade-offs:} While accuracy improves, both advanced models demand significantly more computational power (61.2–61.3 GFLOPs).
\end{itemize}
T-YOLOv5 offers the best balance—strong performance across most classes without the minor inconsistencies of the CBAM variant. However, for scenarios requiring maximum accuracy in complex scenes, T-YOLOv5 with CBAM may be preferred. The choice depends on the task’s resource constraints and detection complexity.

\subsection{Further Tests \& Frame Analysis}
\subsubsection{Sudden Movements}

\begin{figure}[H]
    \centering
    \includegraphics[width=0.5\textwidth]{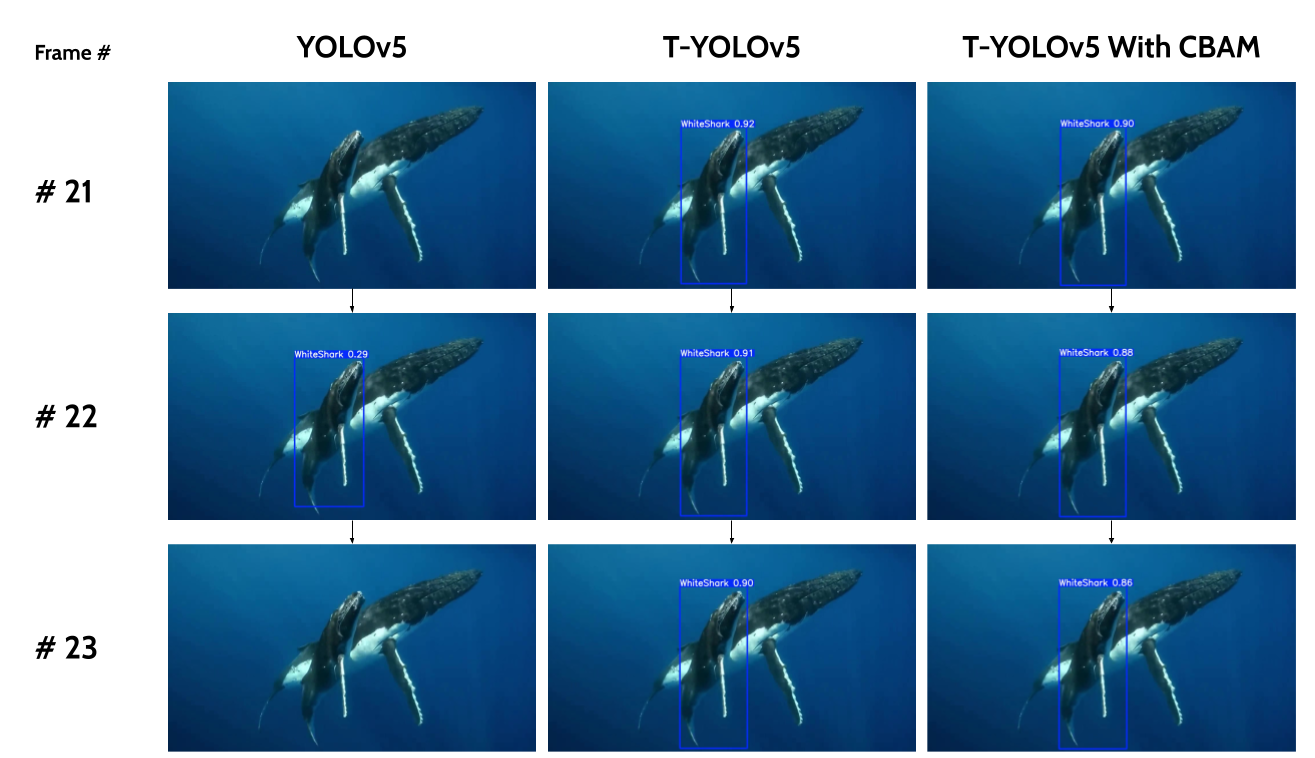}
    \caption{Frame Analysis Of White Shark}
    \label{fig:dataset_samples}
\end{figure}

In the sequence of frames (\#21–\#23) featuring a whale, the standard YOLOv5 struggles with maintaining consistent confidence levels when the object undergoes sudden shifts in movement or shape changes. The standard YOLOv5 exhibits a decrease in detection reliability and lower confidence scores, indicative of its limitations in temporal coherence and adaptability to abrupt changes.

On the other hand, T-YOLOv5 and T-YOLOv5 with CBAM maintain higher confidence levels throughout the sequence. The temporal modeling of T-YOLOv5 allows it to effectively track and identify the whale even when movement patterns change rapidly. It achieves the highest confidence scores compared to the CBAM variant. This is primarily because while T-YOLOv5 with CBAM integrates attention mechanisms that enhance spatial feature selection, it slightly reduces the temporal adaptability by prioritizing intricate details. Therefore, the standard T-YOLOv5, with its simpler architecture, excels by retaining strong temporal awareness without over-focusing on finer features.

\subsubsection{Partially Occluded Objects}

\begin{figure}[H]
    \centering
    \includegraphics[width=0.5\textwidth]{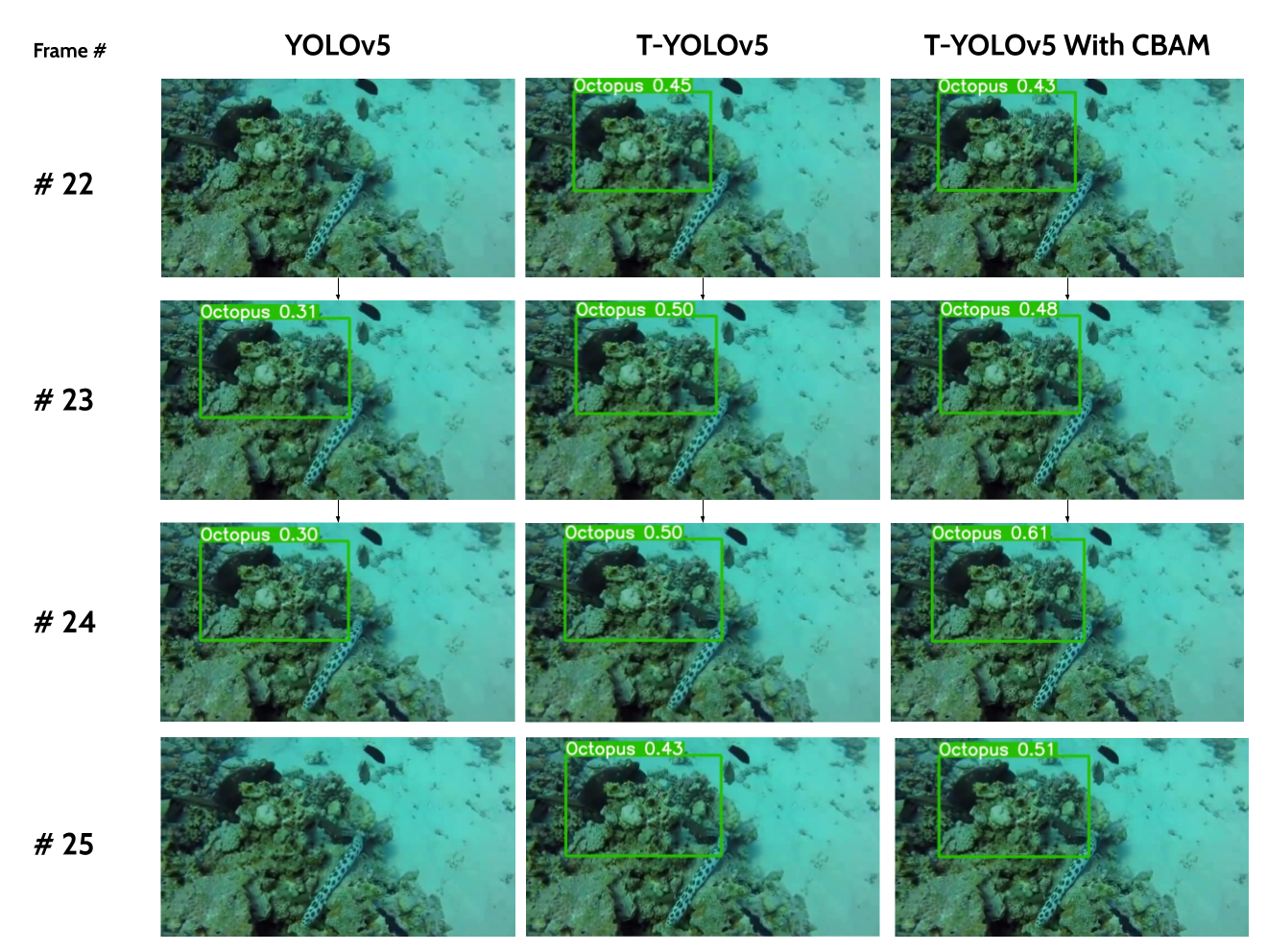}
    \caption{Frame Analysis Of Octopus}
    \label{fig:dataset_samples}
\end{figure}

In the octopus detection frames (\#22–\#25), standard YOLOv5 shows notable limitations when the object is partially obscured by surrounding elements such as rocks or coral. Its detection confidence is considerably low, fluctuating around 0.3 to 0.4, reflecting its struggle to handle occlusions effectively.

The T-YOLOv5 and T-YOLOv5 with CBAM models demonstrate robustness in this scenario. The temporal modeling in T-YOLOv5 helps maintain consistent object tracking and detection, and the CBAM-enhanced version showcases even higher confidence (e.g., scores around 0.5 - 0.6). This superior performance by T-YOLOv5 with CBAM can be attributed to the CBAM's ability to prioritize significant spatial and channel information, focusing on areas of the frame where the object partially reappears, thereby improving detection confidence despite occlusions.

\subsubsection{Visible Gradually Moving Objects}

\begin{figure}[H]
    \centering
    \includegraphics[width=0.5\textwidth]{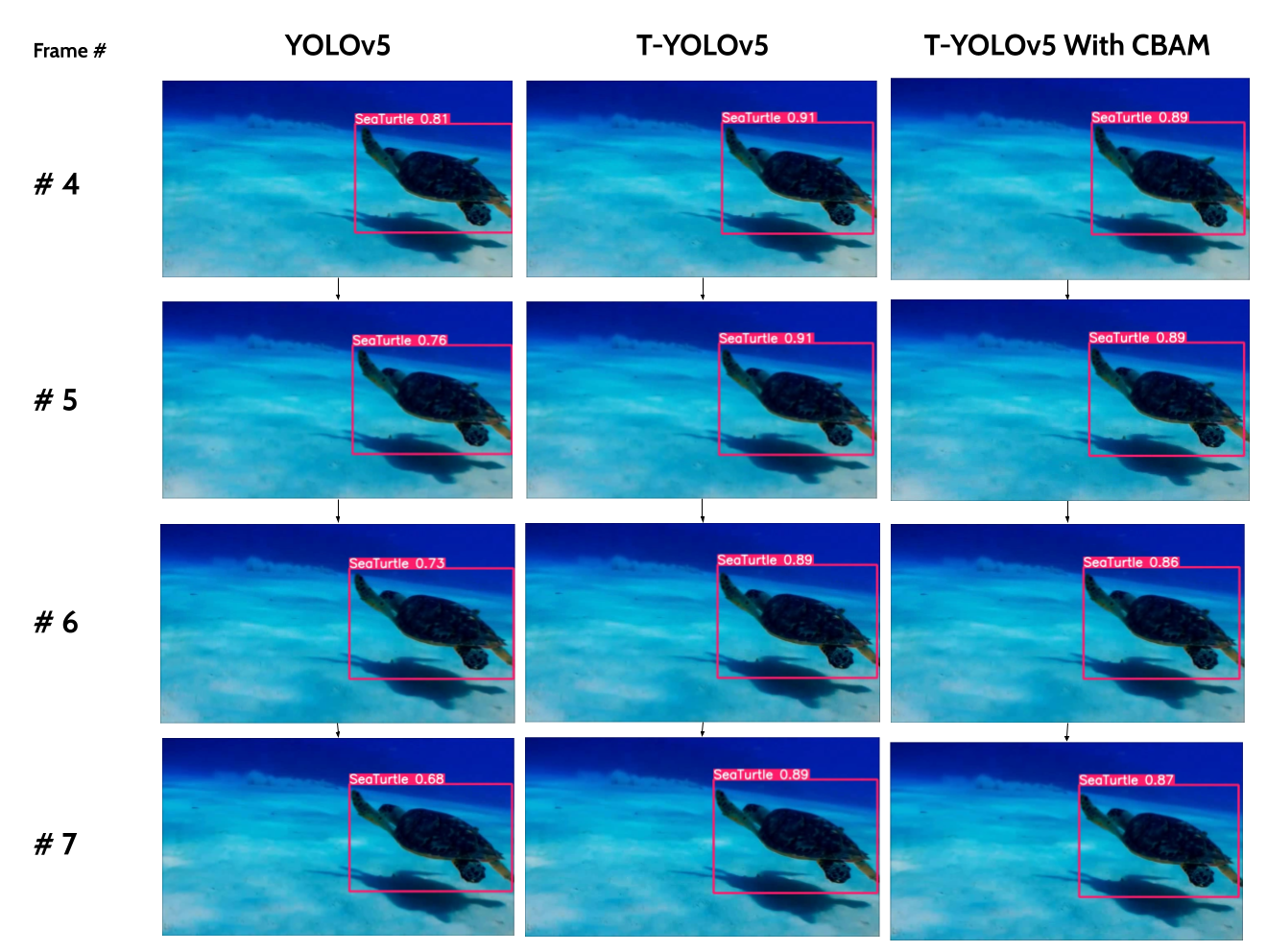}
    \caption{Frame Analysis Of Sea Turtle}
    \label{fig:dataset_samples}
\end{figure}

The final frame sequence (\#4–\#7) shows a gradually moving sea turtle. All three models—YOLOv5, T-YOLOv5, and T-YOLOv5 with CBAM—successfully detect the turtle, but key differences emerge. Standard YOLOv5 shows declining confidence across frames (from 0.81 to 0.68), indicating difficulty maintaining detection stability during subtle movements.

In contrast, T-YOLOv5 and its CBAM variant sustain high-confidence detections (0.8–0.9). T-YOLOv5 demonstrates slightly greater stability, likely due to its simpler architecture focused on temporal coherence without added complexity. The CBAM version remains consistent but occasionally emphasizes spatial details that aren’t essential in clear, gradual-motion scenarios, resulting in slightly lower confidence.

\subsubsection{Deductions From Above Scenarios}

\begin{itemize}
    \item \textbf{Sudden movements / shape changes (e.g., whale):} T-YOLOv5 is preferred for its strong temporal modeling and stability.
    \item \textbf{Partially occluded objects (e.g., octopus):} T-YOLOv5 with CBAM excels, using attention to focus on key areas despite occlusions.
    \item \textbf{Gradual motion (e.g., sea turtle):} Both temporal models outperform YOLOv5, with T-YOLOv5 having a slight edge due to consistent, high-confidence tracking.
\end{itemize}

\section{Conclusions}
This study compared standard YOLOv5, T-YOLOv5, and T-YOLOv5 with CBAM for detecting marine animals under conditions like sudden motion, occlusion, and gradual movement. While YOLOv5 showed basic detection ability, it struggled in dynamic and obstructed settings.

T-YOLOv5 was the most effective, maintaining high confidence during abrupt movements due to strong temporal coherence—key for real-time wildlife tracking. T-YOLOv5 with CBAM improved detection of partially occluded objects but sometimes sacrificed temporal accuracy by overemphasizing spatial features, partially supporting our hypotheses (H1, H2).

Both temporal models outperform YOLOv5 on gradual motion, validating the value of temporal modeling. However, hypothesis H3—expecting better frame-to-frame differentiation and fewer false positives—was only partially met.

Overall, T-YOLOv5 offered strong performance without the added complexity of CBAM. Future research could explore integrating lightweight attention modules, refining temporal modeling for motion blur, or using transformer-based architectures. Another promising direction involves deploying these models on underwater robots to test real-time detection in natural habitats, bridging the gap between lab performance and field application in marine conservation.

\nocite{*}

\end{document}